\documentclass[runningheads]{llncs}
\usepackage{graphicx}
\usepackage{amssymb}
\usepackage{graphicx}

\usepackage{hyperref}       
\usepackage{url}            
\usepackage{booktabs}       
\usepackage{amsfonts}       
\usepackage{nicefrac}       
\usepackage{microtype}      
\usepackage{xcolor}         
\usepackage{amsmath}
\usepackage{subfigure}
\usepackage{algorithm}
\usepackage{algorithmicx}
\usepackage{algpseudocode}

\usepackage{textcomp}
\usepackage{wrapfig}
\usepackage{makecell}

\usepackage{multicol}

\begin{document}
\title{Federated Learning with Limited Node Labels}
%
%
\author{Bisheng Tang\inst{1,2,3}\orcidID{0000-0001-6849-3233} \and
Xiaojun Chen\inst{1,2,3}\orcidID{0000-0003-0362-847X}\thanks{Corresponding author.} \and
Shaopu Wang\inst{1,2,3}\orcidID{0000-0002-8873-2948} \and
Yuexin Xuan\inst{1,2,3}\orcidID{0000-0001-7887-2309} \and
Zhendong Zhao\inst{1,2,3}\orcidID{0000-0003-0003-019X}} 
\authorrunning{B. Tang et al.}
%
\institute{Institute of Information Engineering, Chinese Academy of Sciences, Beijing, China\and
School of Cyber Security, University of Chinese Academy of Sciences, Beijing, China\and
Key Laboratory of Cyberspace Security Defense}
%
\maketitle              
\begin{abstract}
Subgraph federated learning (SFL) is a research methodology that has gained significant attention for its potential to handle distributed graph-structured data. In SFL, the local model comprises graph neural networks (GNNs) with a partial graph structure. However, some SFL models have overlooked the significance of missing cross-subgraph edges, which can lead to local GNNs being unable to message-pass global representations to other parties' GNNs. Moreover, existing SFL models require substantial labeled data, which limits their practical applications. To overcome these limitations, we present a novel SFL framework called FedMpa that aims to learn cross-subgraph node representations. FedMpa first trains a multilayer perceptron (MLP) model using a small amount of data and then propagates the federated feature to the local structures. To further improve the embedding representation of nodes with local subgraphs, we introduce the FedMpae method, which reconstructs the local graph structure with an innovation view that applies pooling operation to form super-nodes. Our extensive experiments on six graph datasets demonstrate that FedMpa is highly effective in node classification. Furthermore, our ablation experiments verify the effectiveness of FedMpa.
\keywords{Subgraph  \and Federated learning \and Scarce sample \and Node classification \and Feature learning \and Structure learning.}
\end{abstract}
\section{Introduction}
Recently, graph-structured data mining has been gaining increasing attention \cite{perozzi2014deepwalk,feng2020graph,yang2021extract,tian2020makes,yu2020graph,wu2020graph,tishby2000information,sun2019infograph}, since GNNs have appeared as highly effective methods to learn graph-structured data embeddings. However, in reality, graph-structured data owners often operate in a distributed environment, unable to observe the global distribution of the data due to their locally biased graphs. Meanwhile, local data owners cannot access data from other owners due to the privacy policy, which seriously affects the expression ability of the model. For instance, different banks forbid providing customer and finance data, and e-commerce platforms cannot share their promotion data with other competitive platforms. These realistic scenarios prevent the data owner from learning powerful models from external knowledge. Recently, federated learning has gained considerable attraction since FedAvg \cite{mcmahan2017communication} has been used to address the cross-silo information problem with privacy protection.

Federated Learning (FL) \cite{mcmahan2017communication,li2020federatedover,karimireddy2020scaffold,arivazhagan2019federated,liang2020think}, a distributed learning framework based on privacy protection, has been proven advantageous in tackling cross-silo information problems without sharing raw data. However, when leveraging FL to learn graph-structured data,  a discrepancy appears. SFL not only needs to aggregate the model parameters as FL does but also needs to pass messages with subgraph structures. The dilemma of subgraph structure being physically isolated has drawn much research attention in recent years \cite{zhang2021subgraph}. Meanwhile, building effective connections (e.g., virtual edges) to form a global graph has gradually become the chief solution, including cross-client information reconstruction \cite{zhang2021subgraph,peng2022fedni} and overlapping instance alignments \cite{cheung2021fedsgc,chen2020vertically}. 

To build effective connection, graph message-passing of the SFL needs to acquire or learn external knowledge (e.g., structures) from other participants. However, existing SFL model learning with other party knowledge requires massive labeled data for high performance, which is often difficult to obtain in real-world scenarios. Meanwhile, learning with massive labeled data generally costs lots of online calculation time in the training phase. For example, FedSage+ \cite{zhang2021subgraph} trains the local graph models by simulating the missing neighbor generation. However, this training process requires massive data and communication costs to train GNNs in the SFL setting, as the FedSage training-validation-testing ratio is 60\%-20\%-20\%. When faced with scarce labeled data, FedSage+ performs poorly. Furthermore, FedSage+ also needs to generate some cross-subgraph edges, which may not exist in real-world scenarios and require a heavy computation effort with other parties' knowledge. Since recovering the missing neighbor and missing edges both require a large amount of labeled data, overcoming these two drawbacks can convert to address the following problem: how to conduct a trainable SFL and simultaneously generate effective virtual cross-subgraph edges when samples are scarce?

To address this challenging problem, we propose a new SFL framework, FedMpa, to address the dilemma of clients having limited labeled data. Inspired by FedSage+, we propose FedMpae to learn global information with global structure views. Note that the generated cross-subgraph edge in FedSage+ is virtual links, which connect subgraphs by generating new neighbor nodes from global knowledge. Intuitively, such an operation may be unnecessary due to the graph pooling technology, which can further condense generated nodes into their ego node to form a super-node. With such observation, we address the cross-subgraph edges differently from FedSage+, which repairs them by simulative generating new nodes and edges. Instead, we innovatively learn cross-subgraph edges by directly pooling the local missing structure (nodes together with edges) into a new super-node and then relearning the edge weights of the new graph (for details, see Figure~\ref{figure2}), which may avoid abundant redundant calculation. Our FedMpae also has an advantage in online calculation speed, as we do not need to generate a node feature representation and can naturally fuse the learning process into GNN's training through the norm constraint.

Our \textbf{contributions} are as follows:

(\romannumeral1) We assume a low label rate SFL scenario which is more realistic, and propose a novel SFL framework, FedMpa, to learn subgraph node representations from the local graphs of multiple parties. Our work is a supplement to FedSage in the low label rate condition, which means our work has performed well in more practical settings.

(\romannumeral2) To better capture the global features in the local subgraphs, we propose the FedMpae model, which can adaptively enhance the subgraph structure to improve the calculation of cross-subgraph edges. FedMpae relearns the local structure to pass the global representation more suitably, which is the key pivot to promote local model expressiveness and generalization.

(\romannumeral3) Extensive experiments on six graph datasets demonstrate the effectiveness of the series of the FedMpa models, achieving competitive performance in node classification.


\section{Related Works}
\paragraph{\textbf{Graph Neural Networks.}}
GNNs aim to learn node representation based on graph topology. Recent GNNs improve model structure with the spatial and frequency domain insights, such as GCN \cite{kipf2016semi}, GAT \cite{velivckovic2017graph}, ST-GCN \cite{yu2017spatio}, SGC \cite{wu2019simplifying}, GraphSage \cite{hamilton2017inductive}, APPNP \cite{klicpera2018predict}, GCNII \cite{chen2020simple}, etc. Based on these basic models, graph pooling skills can further generate the holistic graph-level representation of the entire graph through node clustering/drop pooling \cite{su2021hierarchical,liu2023exploring}. The cross-subgraph edges can be linked with graph coarsen \cite{jin2021graph,cai2021graph}, which can locally pool the missing nodes and edges into a new entity. In this paper, we have improved the learning process of Approximate Personalized Propagation of Neural Predictions (APPNP) to achieve the graph federated process.

\paragraph{\textbf{Federated Learning.}}
Due to data privacy concerns, various institutions cannot synergistically train machine learning models, thus discarding the data value between each collaborator. To address this problem, FL has been proposed \cite{mcmahan2017communication,luo2019real} \cite{jallepalli2021federated,long2020federated,long2022federated}. The vanilla FL method, known as FedAvg \cite{mcmahan2017communication}, has drawn much attention for solving the collaborative training problem, as it averages the parameters' gradients with stochastic gradient descent (SGD). The recent FL proposes a meta-learning framework \cite{finn2017model,hospedales2020meta}, which aims to learn a general model wherein data owners can flexibly fine-tune the model to suit various tasks (i.e., local and global). Personalized FL \cite{deng2020adaptive,fallah2020personalized,kulkarni2020survey}, focusing on the local models, primarily aims to learn an augmented model through knowledge sharing. In the distributed subgraph system, FL is a data augmentation learning technique for local models to some extent, which further motivates us to learn local GNN models from a knowledge augmentation perspective.
\paragraph{\textbf{Graph Federated Learning.}}
Recently, researchers have gradually paid attention to graph federated learning and achieved some impressive results \cite{he2021fedgraphnn,wu2021fedgnn,xie2021federated,zhang2021subgraph} \cite{zhu2021shift,lin2022resource,meng2021cross,lei2023federated,baek2023personalized}. \cite{wu2021fedgnn} considers a scenario that assumes subgraphs have overlapped item nodes and distributed user-item edges, which ignores the relationship of cross-subgraph. FedSage \cite{zhang2021subgraph} and FedNI \cite{peng2022fedni} have addressed a distributed subgraph system with missing cross-subgraph edges by introducing missing nodes that implicitly connect the subgraphs to pass the global information. However, we consider the missing nodes and edges as a whole-global node relying on pooling operation, i.e., we cope with the missing link between each subgraph by introducing abstract edges and thus realize knowledge sharing.

In this work, we also consider the commonly studied scenario, as done in the work of FedSage, where a distributed subgraph system with missing cross-subgraph edges exists. Under this scenario, we restructure the abstract edges using the learned federated features, thus having achieved competitive performance in the local task.

\begin{figure}[htbp]
\centering
\includegraphics[width=1.05\columnwidth]{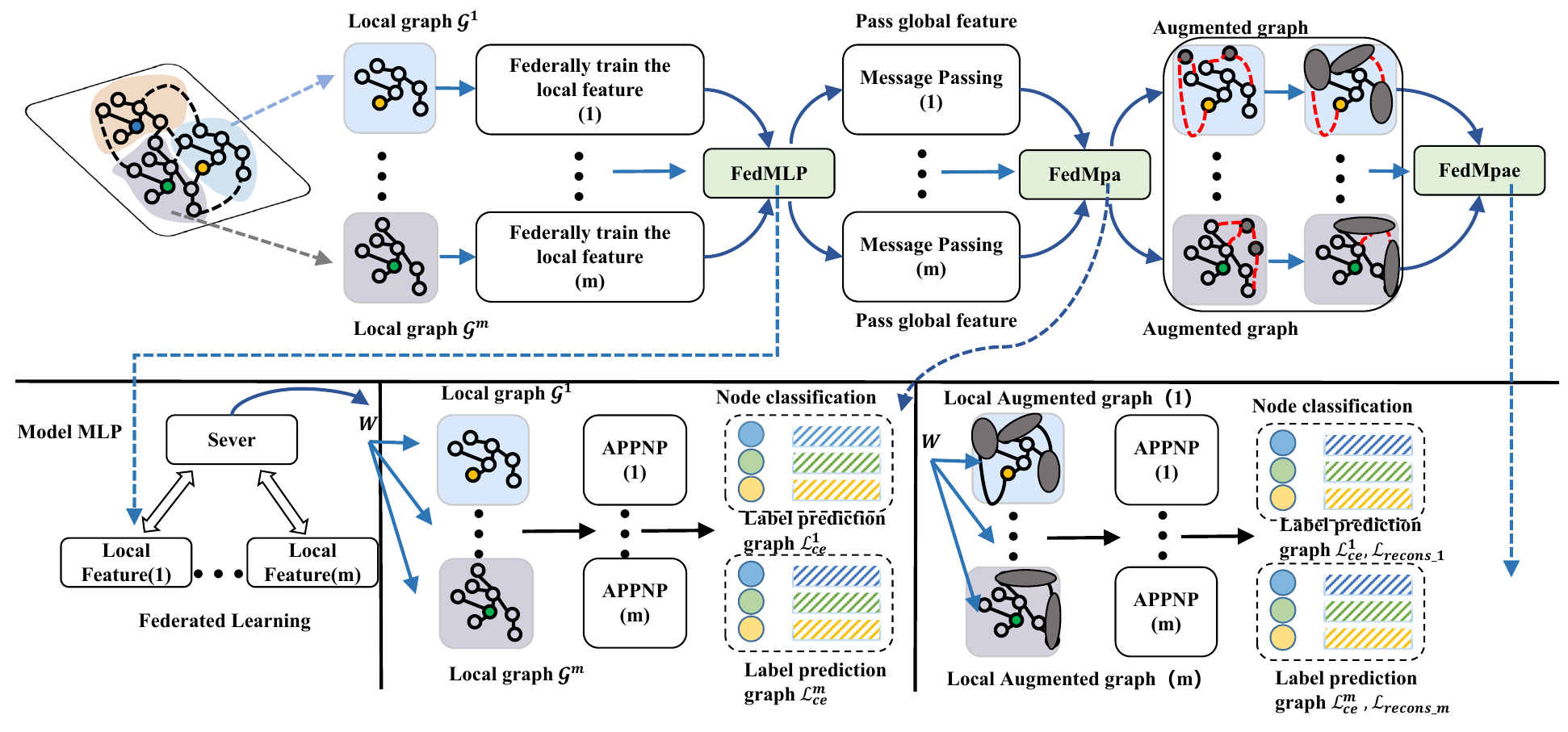} 
\caption{The upper half of the FedMpa framework outlines the process. First, we use FedAvg to train the local feature and set the local model as MLP. After obtaining the FedMLP model, we gain the global information of each local graph. With the learned feature, we can conduct local message passing with approximate PageRank, which we term FedMpa. To handle the cross-subgraph edges, we use graph augmentation to reconstruct the discarded edges, as shown in Figure~\ref{figure2} right part, which we term FedMpae. The bottom half of the framework includes the FedMLP, FedMpa, and FedMpae modules, which detail the learning process and node classification task, including the cross entropy loss $\mathcal{L}_{ce}$ and the local model. After FedMLP has finished, the parameter $W$ is separately transmitted to FedMpa and FedMpae.}
\label{figure1}
\end{figure}
 
\section{Our Proposed Methods: FedMpa and FedMpae}
In this section, we first provide an overview of the related notations of GNNs and FL. We then introduce our proposed FedMpa, which first utilizes FedMLP to learn global feature embeddings. These embeddings are then diffused to each node using the diffusion model, which we term the FedMpa process. Finally, we propose FedMpae, a method for augmenting the local feature representation to improve local model performance. We have shown the relevant learning process in Figure \ref{figure1} and learn details in Algorithm \ref{alg1} and \ref{alg2}.

\paragraph{\textbf{Notations.}}Given a graph $G=(A, V, X)$ in server $\mathbb{S}$, where $A \in \mathbb{R}^{N \times N}, V \in \mathbb{N}^{N}, X \in \mathbb{R}^{N \times d_{0}}$ denote node adjacency matrix, node, and node feature respectively. In the federated learning setting, the data owner (i.e., client $\{\mathbb{C}_{i}| i \in [M]\}$) has a subgraph $G_{i}=(A_{i}, V_{i}, X_{i})$. The graph $G$ satisfy $V=V_{1}\cup V_{2}\cup \cdots \cup V_{M}$, and all the node label $\{y_{v}|v \in V\}$. The local classifier learns local node representation $R_{i}$ based on local graph mapping function $f_{i}:(A_{i}, V_{i}, X_{i}) \to R_{i}$, and subsequently applies the representation $R_{i}$ in downstream task learning(i.e., node classification here). Federated learning settings exchange the learnable parameters $W_{i}$ in separate clients with non-i.i.d data to obtain a global parameter-shared network, whose parameters are calculated as $W:= \sum_{i=0}^{M}{{\lambda}_{i} W_{i}}$, where $\lambda_{i}$ denotes the weight of $W_{i}$.

\paragraph{\textbf{Motivations.}}Consider the matrix equation $\mathbf{\Pi}=P \psi(\mathbf{\Pi})$, where $P$ represents the probability transition matrix, $\psi:\mathbb{R}^{N \times d_{l}}\to \mathbb{R}^{N \times d_{l+1}}$ denotes the mapping function, and $\mathbf{\Pi}$ denotes the status matrix. Generally, the GCNs' training process can be formalized as $\mathbf{\Pi}=P \psi(\mathbf{\Pi})$. However, this training process is often limited by the $\psi$ function and is difficult to converge without label guiding. Instead, the equation $\mathbf{\Pi}=P \mathbf{\Pi}$ can be viewed as a Markov process, which directly produces distinguishable $\mathbf{\Pi}_i$ vectors without labels guiding upon the Markov process converges. We thus can train a simple model to classify converged $\mathbf{\Pi}_i$ vectors using only a few labeled samples. APPNP's diffusion process belongs to $\mathbf{\Pi}=P \mathbf{\Pi}$ iteration form, which benefits SFL with small label rates.

Inspired by the mentioned analysis and to build a federated process, we innovatively decoupled the APPNP into two stages. The first stage involves learning a low-dimensional embedding, which we formalize as follows:
\begin{equation}
R^{(0)}=\mathbf{MLP}(X),
\end{equation} 
where $\mathbf{MLP}$ is a multi-layer perceptron. With the learned feature representation $R^{(0)}$, APPNP then approximates the propagation of the low-dimension feature with personal PageRank, which expresses as follows:
\begin{equation}
R^{(k)}=(1-\alpha) \tilde{A} R^{(k-1)}+\alpha R^{(0)}, \quad \alpha \in (0,1],
\end{equation} 
where $\alpha = 0.1$ is the teleport (or restart) probability, $\tilde{A} = \hat{D}^{-1/2}\hat{A}\hat{D}^{-1/2}$ is the symmetrically normalized adjacency matrix with self-loops and $\hat{D}$ is the diagonal degree matrix $(D+I)$ corresponding to $\hat{A} = A+I$. APPNP then utilizes the $softmax$ function to predict the node label. The final output is
\begin{equation}
R = \mathbf{softmax}(R^{(k)}).
\end{equation} 
In general, APPNP serially trains the two stages in one epoch, which directly couples with calculation in the process of feature transformation. Scalable APPNP appropriately separates the two learning stages and is well suited for SFL tasks. We will explain this process in the next section.
\subsection{Subgraph Federated Feature Learning: FedMLP}
As mentioned above, we first train the MLP model to obtain $R^{(0)}$. The second stage involves passing node information through the diffusion model. Since the first stage aims to learn a low-dimension embedding without considering the graph structure, we can flexibly train the MLP model in different ways, e.g., distributed or centralized training. In the FL setting, traditional methods such as FedAvg can address the non-i.i.d feature label problem (with i.i.d training samples), which makes it easy to federally train an MLP model with different parties' graph features, which we formalize as follows:
\begin{equation}
R_{i}^{(0)}=\mathbf{MLP}(X_{i}), \quad i \in [M] \quad and \quad X_i \sim \textbf{i.i.d}(X).
\end{equation} 
We calculate the parameters $W$ of $MLP(\cdot)$ models by performing federally learning between each client $\mathbb{C}_{i}$, using the FedAvg algorithm as follows:
\begin{equation}
\begin{split}
W=\mathbf{fedAvg}(W)= \sum_{i=0}^{M}{\frac{1}{|M|} W_{i}} \\
s.t. \quad W_{i}=\{W_{i}|X_{i}, epoch\}, 
\end{split}
\end{equation} 
where $\mathbf{fedAvg}(\cdot)$ is used to averagely aggregate the parameters of model $\mathbf{MLP}_{i}(\cdot)$ and $epoch=20$ denotes the communication rounds. We utilize cross-entry loss to optimize subgraph representation learning, and the relevant data split is in Table \ref{tab1}. As the federally aggregated features include global feature information, we can reformulate the federal feature $R_{i}^{(0)}$ as $\hat{R}_{i}^{(0)}=\mathbf{MLP}(X_{i}|W(X_{i}))$, where $W=W_{1}=W_{2} \cdots W_{M}$, and we term this MLP paradigm as FedMLP.

\subsection{Subgraph Federated Structure Learning}
After considering the non-i.i.d problem in the feature aspect, we must also consider the non-i.i.d issue in the subgraph structure. Adding global information affects the status of the subgraph's nodes $V_{i}$ and edges $E_{i}$, demonstrating that the existing subgraph structure is inadequate for absorbing improved global knowledge.

\paragraph{\textbf{Global Feature Learning with Local Structure: FedMpa.}} We propose FedMpa, a method with global features to train a subgraph model in the local client $\mathbb{C}_{i}$, which aims to learn node representation federally. After the FedMLP process is over, we can further learn feature representation with approximate personalized PageRank in the local subgraph, which we formalize as follows:
\begin{equation}
\hat{R}^{(k)}=(1-\alpha) \tilde{A} \hat{R}^{(k-1)}+\alpha \hat{R}^{(0)}.
\end{equation} 
FedMpa learns $\hat{R}^{(k)}$ through a global feature in FL. The subgraph node classification utilizes $\hat{R}_{i}^{(k)}$ (with local adjacency matrix $A_{i}$) to calculate the softmax probability representation of $\{v\in V_{i}|i\in[M]\}$.

\paragraph{\textbf{Global Structure Learning with Global Feature: FedMpae.}} Although FedMpa has learned the global representation in a FL paradigm, it is still worth noting that local structure $A_i$ can be better adjusted using a more appropriate method. To this end, we propose FedMpae, which uses graph autoencoders, a self-supervised learning paradigm, to reconstruct the subgraph structure. The graph autoencoder first learns nodes representation $Z$ as:
\begin{equation}
Z=\mathbf{Encoder}(X)=\mathbf{GNN}(A,X).
\end{equation} 
The GNN models are usually employed as the encoder to transform the input feature $X$ into a graph-structured node feature $Z$. We then use the $Z$ to non-probabilistically calculate the reconstruction matrix $\bar{A}=\mathbf{Decoder}(Z)=\sigma(Z\cdot Z^{T})$, with $\sigma$ being the sigmoid function.

\begin{figure}[htbp]
\centering
\begin{minipage}[c]{0.4\textwidth}
\centering
\includegraphics[width=\columnwidth]{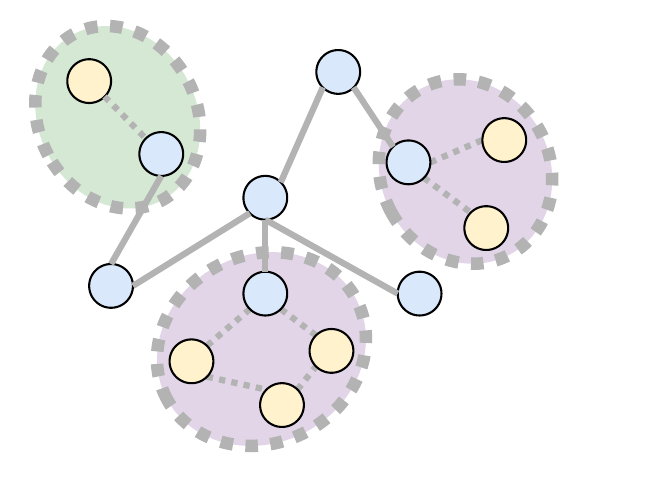}\label{figure2_1}
\end{minipage}
\hspace{0.02\textwidth}
\begin{minipage}[c]{0.4\textwidth}
\centering
\includegraphics[width=\columnwidth]{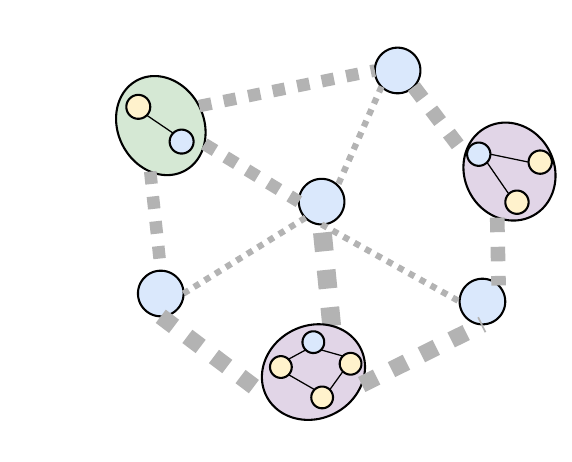}\label{figure2_2}
\end{minipage}
\caption{Left: Generate the missing neighbor node in subgraph with global feature. Right: Repair the missing link or link weight in subgraph with global feature. These two figures have presented two approaches for repairing the subgraph with a global view. On the left, the orange nodes and corresponding dashed edges do not exist in the local client, so costly calculations are required to simulate the generation of the missing nodes and edges across the subgraph. On the right, instead, we combine the missing neighbors (i.e., orange nodes) with edges to form a new single entity (i.e., a new super-node), whose augmented node feature needs to learn from the federated paradigm, and the corresponding edge weights (dashed lines) are also made learnable.}
\label{figure2}
\end{figure}

\renewcommand{\algorithmicrequire}{\textbf{Input:}}
\begin{algorithm}
\caption{Training algorithm of FedMpa on the client side}
\label{alg:algorithm}

\leftline{\textbf{Input}: i-th Local Adjacency Matrix $A_{i}$, Feature Matrix $X_{i}$, and label $Y_{i}$.}
\leftline{\textbf{Parameter}:$W_{i}$, where $i \in [M]$.}
\leftline{\textbf{Output}: Trained client-side model weights $W_{i}$, $W_{FedMpa\_i}$, $W_{FedMpae\_i}$.}
\leftline{\textbf{Client executes:}}
\begin{algorithmic}[1] 

\For {each client $i \in [M]$ in parallel}
\For {client round $1, 2, \cdots, epoch$}  \Comment{FedMLP:}
\State $R_{i}^{(0)}=MLP(X_{i})$ \Comment{The federated learning process FedMLP}
\State $W_{i} \leftarrow \mathbf{ServerUpdate}(\nabla MLP(X_{i}))$ \Comment{Use cross-entropy loss}
\EndFor
\EndFor
\For{each client $i \in [M]$}         \Comment{FedMpa:}
\State $W_{FedMpa\_i} \leftarrow W_{i}$
\For {$k=1, 2, \cdots, l$ round}      
\State $\hat{R}^{(k)}_{i}\leftarrow \mathbf{MessageDiffusion}(A_{i},R_{i}^{(0)},W_{FedMpa\_i})$ \Comment{GNN diffusion training process: model APPNP}
\State $W_{FedMpa\_i} \leftarrow \mathbf{BackwardUpdate}(\nabla_{W_{FedMpa\_i}} CrossEntropy(\hat{R}^{(k)}_{i},Y_{i})$  \Comment{ Gradient backward propagation}
\EndFor
\EndFor
\For{each client $i \in [M]$}      \Comment{FedMpae:}
\State $W_{FedMpae\_i} \leftarrow W_{i}$
\For {$k=1, 2, \cdots, l$ round}
\State $\check{R}^{(k)}_{i}\leftarrow \mathbf{MessageDiffusion}(\check{A_{i}},R_{i}^{(0)},W_{FedMpae\_i})$ \Comment{APPNP Encoder: GNN diffusion training process}
\State $\bar{A}\leftarrow \mathbf{Decoder}(\check{R}^{(k)}_{i})$
\State $\mathcal{L}_{recons\_i}=a\cdot ||A_{i}-\check{A}_{i}||_{2}+b\cdot ||\bar{A}_{i}-\check{A}_{i}||_{2}$ \Comment{Calculate reconstruction loss}
\State $\mathcal{L}_{FedMpae\_i}=\beta \cdot CrossEntropy(\check{R}^{(k)}_{i}, Y_{i})+\gamma \cdot \mathcal{L}_{recons\_i}$
\State $W_{FedMpae\_i} \leftarrow \mathbf{BackwardUpdate}(\nabla_{W_{FedMpae\_i}} \mathcal{L}_{FedMpae\_i})$  \Comment{FedMpae gradient backward propagation}
\EndFor
\EndFor
\State \textbf{return} 
\end{algorithmic}
\label{alg1}
\end{algorithm}

\begin{algorithm}[htbp]

\caption{ Training algorithm of FedMpa on the server side.}
\leftline{\textbf{SeverUpdate($\nabla_i$):}} 
\begin{algorithmic}[1] 
\For {$i \in [M]$}
\State $\nabla=\sum_{i=0}^{[M]}\lambda_{i}\cdot \nabla_{i}$ \Comment{Use fedAvg to aggregate the gradient collected from each client}
\EndFor

\State \textbf{return} $\nabla$
\end{algorithmic}\label{alg2}
\end{algorithm}

\textbf{Method innovation:} As shown in Figure~\ref{figure2}, there are two ways to mend the subgraph in global views. The first approach relates to FedSage, which simulates generating the missing cross-subgraph edges and nodes. The second solution we proposed involves graph pooling operation \cite{jin2021graph,cai2021graph}, which needs to redefine the weights of the edges. Specifically, we relearn an adjacency matrix $\check{A}$ with super-nodes representing the missing nodes and edges during the training process (we assume such process: $A \in \mathbb{R}^{N \times N}$ $\to$ generate m new nodes $A_{new} \in \mathbb{R}^{(N+m) \times (N+m)}$ $\to$ node pooling $\check{A} \in \mathbb{R}^{N \times N}$ ). Consequently, the representation $Z$ can be formulated as $Z=\mathbf{GNN}(\check{A}, X)$. We define a reconstruction loss to optimize $\check{A}$:
\begin{equation}
\mathcal{L}_{recons}=a\cdot||A-\check{A}||_{2}+b\cdot||\bar{A}-\check{A}||_{2}.
\end{equation} 
If we set a trick with $a=b=1$, then $\mathcal{L}_{recons}=||A-\check{A}||_{2}+||\bar{A}-\check{A}||_{2}\geq ||A-\bar{A}||_{2}$. Easy forward, we only have to optimize the lower bound of $||A-\bar{A}||_{2}$. Consequently, $Z$ can return to $\mathbf{GNN}(A, X)$ with the help of the trigonometric inequality. We use the Mean Squared Error (MSE) loss to calculate the $l_{2}$-norm in our experiments:
\begin{equation}
\mathcal{L}_{recons}=\mathbf{MSE}(A,\bar{A}), \quad with \quad Z=\mathbf{GNN}(A,X).
\end{equation} 
By combining the MSE loss with GNN cross-entropy loss ($\mathcal{L}=\mathbf{CE}(R,y_{v})$), we can obtain the $\mathbb{C}_{i}$ FedMpae loss as follows:
\begin{equation}
\begin{split}
\mathcal{L}_{fedMpae\_i}&=\beta \cdot \mathcal{L}_{i}+\gamma \cdot \mathcal{L}_{recons\_i}\\
&=\beta \cdot \mathbf{CE}(\hat{R}_{i},y_{v})+\gamma \cdot (a\cdot||A-\check{A}||_{2}+b\cdot||\bar{A}-\check{A}||_{2})\\
&\geq\beta \cdot \mathbf{CE}(\hat{R}_{i},y_{v})+\gamma \cdot \mathbf{MSE}(A_{i},\bar{A}_{i}),
\end{split}
\end{equation} 
where $\beta = 1$ and $\gamma = 1$ in experiments are the hyper-parameters for $\mathcal{L}_{i}$ and $\mathcal{L}_{recons\_i}$ respectively. FedMpae seeks to propagate the global feature with the reconstructed structure, which has improved the local model's classification performance in our experiments.

\section{Experiments}
In this section, we evaluate the performance of our FedMpa framework in comparison to the FedSage model on node classification benchmarks. We then conduct ablation experiments, including dropout rate, learning rate, hidden dimension, loss hyper-parameters, and label rate, to verify the effectiveness of our FedMpa framework. Additionally, we demonstrate the superiority of FedMpa in online calculations. We aim to answer three questions as follows. \textbf{Q1:} Can FedMpa achieve competitive performance than FL and GFL models? \textbf{Q2:} How do the hyper-parameters affect FedMpa’s training? \textbf{Q3:} Why can FedMpa complement with FedSage in the low label rate? 

\subsection{Settings}

\paragraph{\textbf{Datasets.}} 
We evaluate the effectiveness of federated learning for node classification on six datasets: Cora, Citeseer, Amazon Electronics Computers (Computer), Amazon Electronics Photo (Photo), Coauthor Microsoft Academic Computing Science (Coauthor-cs), and Coauthor Microsoft Academic Physics (Coauthor-phy). The relevant statistical data is shown in Table \ref{tab1}.

\begin{table}[htbp]
  \centering
  \caption{Statistical information of the datasets.}\label{tab1}
  \scalebox{1}{
   \renewcommand\arraystretch{1.5}
  \begin{tabular}{cccccc}
  \hline
    \textbf{Dataset} & \textbf{\#Nodes} & \textbf{\#Edges} & \textbf{\#Classes} & \textbf{\#Features} & \textbf{\#Train/Valid/Test} \\
  \hline
    Cora      & 2,708 & 5,429 & 7 & 1,433 & 1\%/20\%/20\%\\
    Citeseer  & 3,312  & 4,732 & 6 & 3,703 & 1\%/20\%/20\%\\
\hline
    Computer    & 13,381 &   245,778 & 10 & 767 & 1\%/20\%/20\%\\
    Photo    & 7,487 &   119,043 & 8 & 745 & 1\%/20\%/20\%\\
    Coauthor-cs    & 18,333 &   182,121 & 15 & 6805 & 1\%/20\%/20\%\\
    Coauthor-phy    & 34,493 &   530,417 & 5 & 8415 & 1\%/20\%/20\%\\
   \hline
  \end{tabular}
  }
\end{table}
\paragraph{\textbf{Baselines.}}
FedSage is a subgraph federated paradigm that focuses on not only global tasks but also local tasks. We chose FedSage works as our baselines for this work, namely LocSage, FedSage, and FedSage+, corresponding to our work LocMpa, FedMpa, and FedMpae, where LocMpa means training the APPNP model using local features and structure. Since our work considers a scarce labeled data setting, we need to retrain FedSage by changing its code parameters with a small label rate (e.g., 1\%). FedGCN (2-layer GCN) federally aggregates the weight gradient in various parties through the FedAvg algorithm. The FedProx is achieved using the MLP model and the proximal item according to work \cite{li2020federated}. We implement the recent SOTA SFL model FED-PUB \cite{baek2023personalized} according to the paper provided code. 

\paragraph{\textbf{Implementation Details.}}
We have implemented FedMpae based on APPNP's diffusion process and FedMLP, mainly leveraging the TensorFlow and stellargraph lib. We reused the FedSage code and divided the subgraph like the Louvain algorithm to imitate different client subgraphs in reality (i.e., $|V_{i}|=|V|/ M$). We set the FedMLP as a 3-hidden layer network with 64 dimensions, a 0.5 dropout rate, and a 0.01 learning rate. Meanwhile, we have set the FedMpa and FedMpae with 0.5 dropout rate, 0.1 $\alpha$, and 0.01 learning rate. We conduct experiments on two NVIDIA Tesla V100 GPUs. 

\subsection{Results on Node Classification}
Table \ref{tab2} presents the average accuracy of node classification along with standard deviation, indicating the effectiveness of our method in the six datasets. We compare our approach with the framework FedSage. The results show that FedSage/FedSage+ performs inferior to FedMpa/FedMpae when label data is scarce, as fewer samplable neighbors result in poorer generalization. Moreover, our FedMpa constantly outperforms LocMpa, indicating the effectiveness of the learned federated feature. Furthermore, our FedMpae demonstrates advantages in LocMpa and FedMpa across six datasets, proving the effectiveness of augmented graph representation. We also conduct experiments on different parties, presented in Table \ref{tab3}, to further manifest the advantages of FedMpae. Meanwhile, our method has performance improvement over the FedProx, FedGCN, and recently proposed SOTA SFL model FED-PUB. From the data perspective, our FedMpae outperforms FedSage+ with 64.7\%, 33.7\%, 0.7\%, 4.1\%, 20.0\%, and 36.1\% in Cora, Citeseer, Coauthor-phy, Coauthor-cs, Computer, and Photo, respectively. All those mentioned analysis can answer \textbf{Q1}.

\begin{table}[htbp]
  \centering
    \caption{We conducted experiments on six datasets, comparing the results to those of FedSage. Three clients were used to experiment on six datasets of varying graph data scales.}\label{tab2}
  \scalebox{0.9}{
   \renewcommand\arraystretch{1.5}
  \begin{tabular}{c|c|c|c|c|c|c}
    \toprule
   Acc(\%) & \multicolumn{6}{c}{\textbf{M=3}}\\
\cmidrule(l){2-7}
  \textbf{Model} & \multicolumn{1}{c|}{\textbf{Cora}}& \multicolumn{1}{c|}{\textbf{Citeseer}}& \multicolumn{1}{c|}{\textbf{Coauthor-phy}}& \multicolumn{1}{c|}{\textbf{Coauthor-cs}} & \multicolumn{1}{c|}{\textbf{Computer}} & \multicolumn{1}{c}{\textbf{Photo}}\\

\hline
    FedMLP     & 33.37$\pm$3.32& 31.29$\pm$2.59 & 67.44$\pm$3.48 & 40.59$\pm$4.48 & 41.22$\pm$3.42 & 35.92$\pm$9.09\\
    FedProx & 35.96$\pm$6.33 &32.84$\pm$6.20&69.50$\pm$7.56 & 47.94$\pm$5.81 & 41.29$\pm$3.66& 40.14$\pm$7.48\\
    FedGCN     & 47.12$\pm$7.07& 42.98$\pm$2.58 & 88.26$\pm$2.24 & 66.52$\pm$3.97 & 42.42$\pm$2.29 & 57.82$\pm$5.47\\
    FED-PUB   & 50.46$\pm$0.71 & 53.63$\pm$0.90&80.12$\pm$0.68& 50.75$\pm$7.19& 47.86$\pm$0.83 & 43.31$\pm$0.79 \\
\hline
    LocSage     & 47.60$\pm$4.55& 38.24$\pm$2.23 & 90.58$\pm$0.39 & 80.72$\pm$0.61 & 65.37$\pm$1.77 & 69.96$\pm$0.88\\
    FedSage  & 41.51$\pm$2.88  & 43.86$\pm$3.67 & 92.15$\pm$0.72 & 81.57$\pm$1.25 & 62.75$\pm$3.46 &61.57$\pm$7.34\\
    FedSage+    & 40.72$\pm$3.52  & 40.59$\pm$3.75 & 92.12$\pm$0.57  & 82.85$\pm$1.26 & 62.04$\pm$1.82 &63.26$\pm$3.24\\
    \hline
    LocMpa    & 65.67$\pm$6.55& 51.73$\pm$2.82 & 91.45$\pm$1.72 & 84.95$\pm$0.88 & 71.40$\pm$4.05 &84.31$\pm$1.41\\
    FedMpa    & 65.75$\pm$3.85& 53.99$\pm$3.57 & \textbf{92.80$\pm$0.29} & 85.42$\pm$1.29 & 73.53$\pm$6.28 & 83.53$\pm$2.18\\
    FedMpae    & \textbf{67.07$\pm$3.28} &   \textbf{54.27$\pm$2.50}  & \textbf{92.73$\pm$0.49} & \textbf{86.21$\pm$0.86} & \textbf{74.65$\pm$4.61} & \textbf{86.08$\pm$1.70}\\
\bottomrule
  \end{tabular}
  }
\end{table}
\begin{table}[htbp]
\centering
\caption{Experimental results on the number of clients $M=\{3, 5, 10\}$. Various $M$ is to verify the effectiveness of our model towards the different numbers of clients.}\label{tab3}
\scalebox{0.95}{
   \renewcommand\arraystretch{1.5}
\begin{tabular}{c|c|c|c|c|c|c}
\bottomrule
   Acc(\%)& \multicolumn{3}{c}{\textbf{Citeseer}} & \multicolumn{3}{c}{\textbf{Computer}} \\
\cmidrule(l){2-4} \cmidrule(l){5-7}  
    \textbf{Model} & \textbf{M=3} & \textbf{M=5} & \textbf{M=10}  & \textbf{M=3} & \textbf{M=5} & \textbf{M=10} \\
  \hline
    FedMLP   &  31.29$\pm$2.59 & 29.77$\pm$4.17 & 27.90$\pm$4.79 & 41.22$\pm$3.42 & 41.13$\pm$2.08 & 42.12$\pm$2.04\\
    FedProx & 32.84$\pm$6.20 &38.09$\pm$5.28 &27.57$\pm$3.93 & 41.29$\pm$3.66&44.21$\pm$4.04 &45.47$\pm$2.42 \\
    FedGCN &  42.98$\pm$2.58  & 36.59$\pm$4.09 & 32.22$\pm$4.11 & 42.42$\pm$2.29 & 42.07$\pm$3.71 & 40.25$\pm$2.01\\
    FED-PUB & 53.63$\pm$0.90 & 46.28$\pm$3.36&51.57$\pm$5.90 & 47.86$\pm$0.83 &59.07$\pm$0.52 & 66.30$\pm$0.34\\
\hline
    LocSage   & 38.24$\pm$2.23 & 35.41$\pm$2.97 & 45.71$\pm$1.12 & 65.37$\pm$1.77 & 58.69$\pm$1.69 & 53.15$\pm$1.89\\
    FedSage & 43.86$\pm$3.67  & 36.57$\pm$3.90 & 49.03$\pm$3.19 &  62.75$\pm$3.46 & 49.37$\pm$7.09 & 42.69$\pm$7.53\\
    FedSage+    &  40.59$\pm$3.75 & 37.78$\pm$3.83 & 50.20$\pm$1.10 & 62.04$\pm$1.82 & 44.98$\pm$3.48 & 44.11$\pm$10.44\\
    \hline
    LocMpa    &  51.73$\pm$2.82 & 50.20$\pm$4.20 & 59.78$\pm$2.90 & 71.40$\pm$4.05 & 72.72$\pm$1.74 & 71.74$\pm$6.78\\
    FedMpa    & 53.99$\pm$3.57 &   \textbf{51.73$\pm$3.71} & 60.39$\pm$4.93 & 73.53$\pm$6.28 &   \textbf{73.99$\pm$3.03} & 71.32$\pm$7.78\\
    FedMpae    &  \textbf{54.27$\pm$2.50} & \textbf{51.42$\pm$4.42} & \textbf{62.47$\pm$2.54}&  \textbf{74.65$\pm$4.61} & 71.52$\pm$0.57 & \textbf{72.06$\pm$7.99}\\
\bottomrule
  \end{tabular}
}
\end{table}

\subsection{Parameter Sensitivity Analysis}

\begin{figure}[htbp]
\centering
\begin{minipage}[c]{0.3\textwidth}
\centering
\includegraphics[width=\columnwidth]{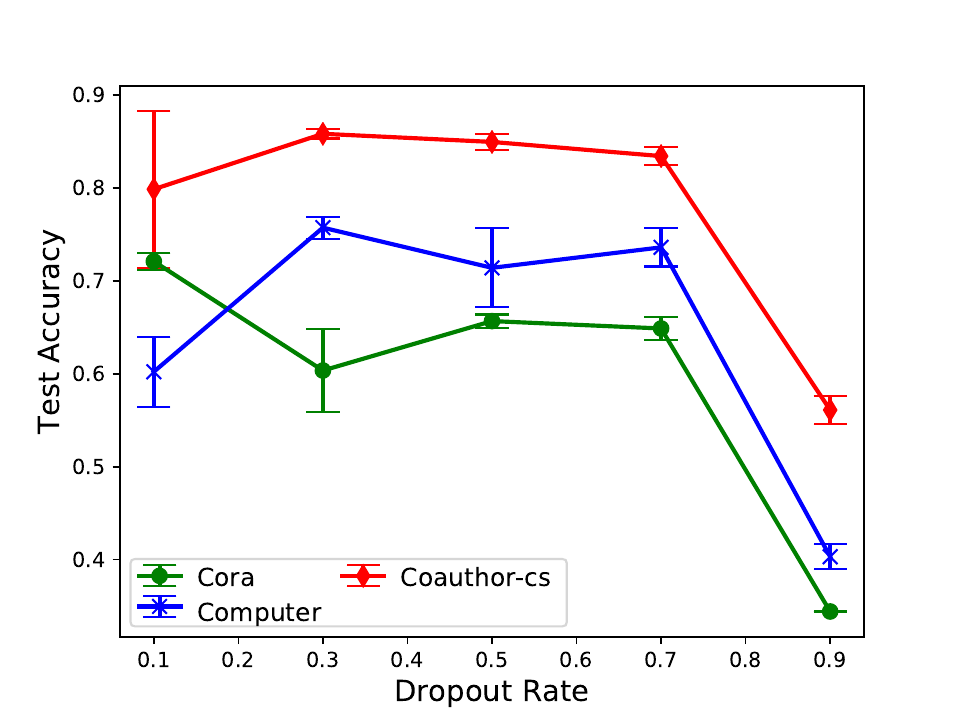}\label{drop_1}
\end{minipage}
\hspace{0.02\textwidth}
\begin{minipage}[c]{0.3\textwidth}
\centering
\includegraphics[width=\columnwidth]{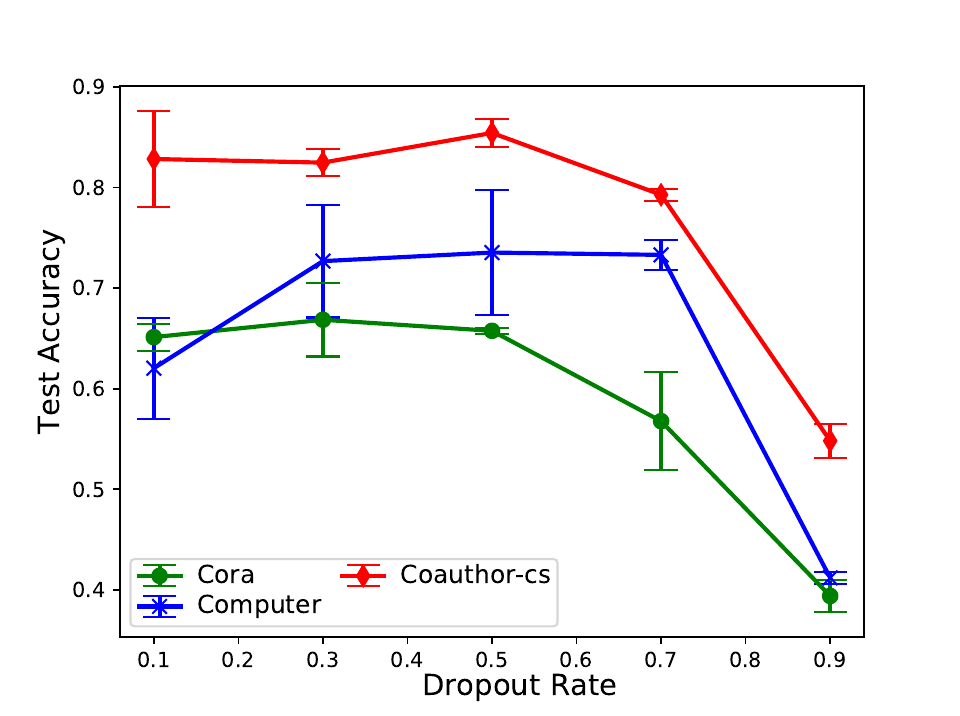}\label{drop_2}
\end{minipage}
\hspace{0.02\textwidth}
\begin{minipage}[c]{0.3\textwidth}
\centering
\includegraphics[width=\columnwidth]{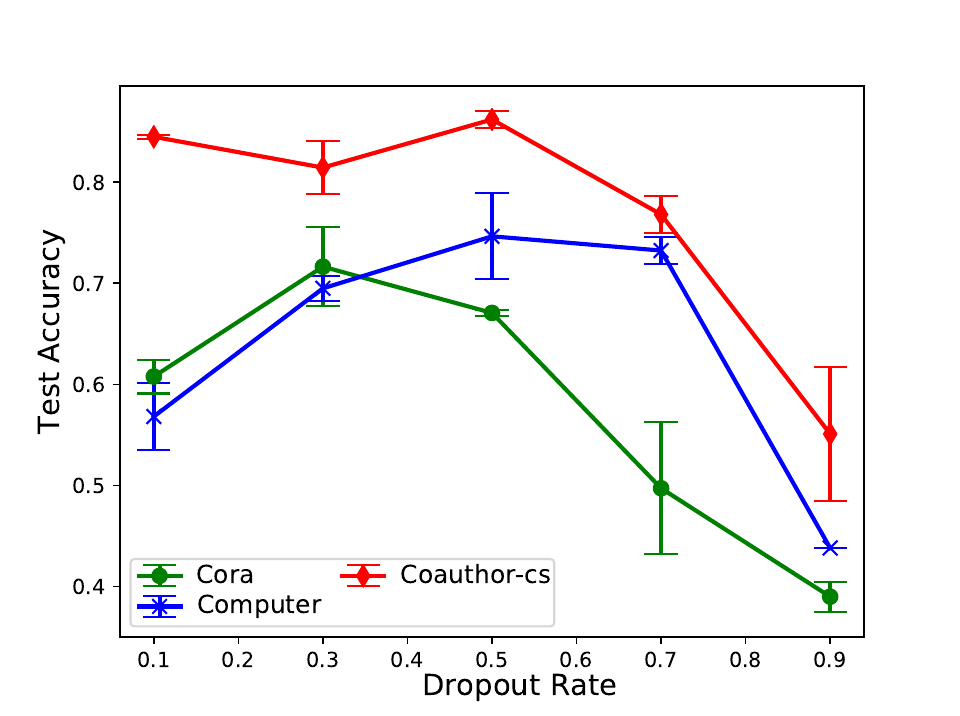}\label{drop_3}
\end{minipage}
\caption{This table illustrates the impact of the dropout rate on node classification for Cora, Coauthor-cs, and Computer with LocMpa, FedMpa, and FedMpae, respectively.}
\label{drop}
\end{figure}

\paragraph{\textbf{Dropout Rate.}}
In this part, we present an analysis of dropout rates ranging from 0.1 to 0.9 on three datasets, namely Cora, Coauthor-cs, and Computer, as shown in Figure~\ref{drop}. Our results indicate that our methods demonstrate almost peak performance with a dropout rate of 0.5. We attribute this to two primary factors: first, a high dropout rate can fail knowledge exchange due to the local model's insufficient training, and second, a low dropout rate can cause overfitting of the local graph model and result in suboptimal local solutions. Based on our findings, we selected a dropout rate of 0.5 as the optimal rate for use in FL settings across all six datasets considered in this study.

\paragraph{\textbf{Hidden Dimension and Learning Rate.}}
\begin{table}[htbp]
\centering
  \caption{Ablation experimental on learning rate (lr) and hidden dimension (h-dim).}\label{tab4}
 \scalebox{0.9}{
   \renewcommand\arraystretch{1.5}
\begin{tabular}{c|c|c|c|c|c|c}
\bottomrule
   Accuracy \%& \multicolumn{3}{c}{\textbf{Cora}} & \multicolumn{3}{c}{\textbf{Computer}} \\
\cmidrule(l){2-4} \cmidrule(l){5-7}  
    \textbf{Model} & \textbf{LocMpa} & \textbf{FedMpa} & \textbf{FedMpae}  & \textbf{LocMpa} & \textbf{FedMpa} & \textbf{FedMpae} \\
  \hline
    lr=0.001   &  63.91$\pm$11.99 & 59.30$\pm$6.25 & 65.93$\pm$1.30 &  78.63$\pm$1.03 & 74.65$\pm$1.41 & 76.14$\pm$3.46\\
     lr=0.005 & 73.39$\pm$0.13 & 67.40$\pm$4.95& 61.79$\pm$1.17 &  76.18$\pm$0.59 & 74.61$\pm$1.16 & 76.16$\pm$0.05 \\
     lr=0.01   &  65.67$\pm$6.55 & 65.75$\pm$3.85 & 67.07$\pm$3.28 &   71.40$\pm$4.05 & 73.53$\pm$6.28 & 74.65$\pm$4.61 \\
     lr=0.02 &  70.81$\pm$0.65 & 67.32$\pm$4.04 & 66.30$\pm$10.16 & 62.49$\pm$13.07  & 70.77$\pm$0.75 & 62.62$\pm$12.85\\
    \hline
   h-dim=16    &  69.43$\pm$0.78 &   58.56$\pm$1.05 & 47.88$\pm$0.78 &   75.07$\pm$0.62 &   75.01$\pm$2.13 & 74.66$\pm$1.19 \\
    h-dim=32    &  66.88$\pm$2.84 &  61.62$\pm$2.23& 62.24$\pm$3.14&  77.81$\pm$0.34 &  74.85$\pm$0.14 & 76.18$\pm$1.00\\
    h-dim=64      &  65.67$\pm$6.55 & 65.75$\pm$3.85 & 67.07$\pm$3.28 &   71.40$\pm$4.05 & 73.53$\pm$6.28 & 74.65$\pm$4.61 \\
    h-dim=128     &  68.42$\pm$2.48 & 64.73$\pm$4.56 & 66.75$\pm$1.68 &  68.14$\pm$1.85 & 59.49$\pm$8.93 & 60.35$\pm$16.71\\
\bottomrule
  \end{tabular}
}
\end{table}
In this part, we investigate the impact of the learning rate and hidden dimension on our model. The results of our experiments, presented in Table \ref{tab4}, reveal that certain learning rates, such as 0.001 and 0.02, exhibit large standard deviation values, which can be detrimental to the training of our model. Therefore, we recommend setting the fixed learning rate parameter to 0.01 for the chosen datasets. As for the effect of hidden dimension, we consider that the optimal value should be determined based on the size of the dataset. In our experiments, we fix the hidden dimension at 64, which we consider the optimal hyper-parameter. This operation simplifies the experimental setup and enables us to focus on other essential aspects of our model.

\begin{figure}[htbp]
\centering
\begin{minipage}[c]{0.45\textwidth}
\centering
\includegraphics[width=\columnwidth]{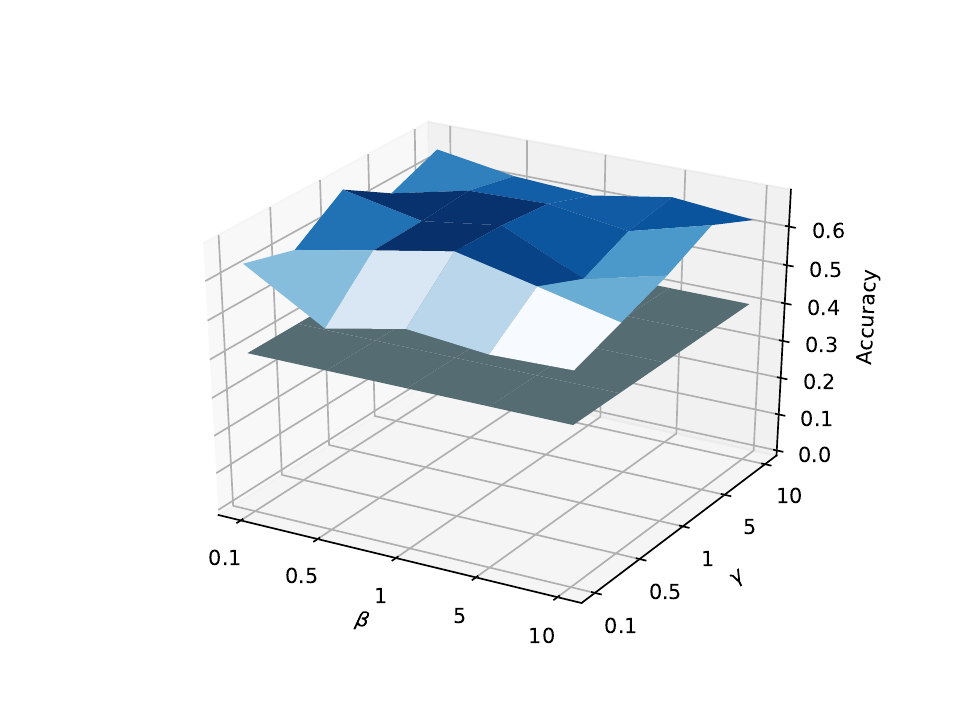}\label{3d_1}
\end{minipage}
\hspace{0.02\textwidth}
\begin{minipage}[c]{0.45\textwidth}
\centering
\includegraphics[width=\columnwidth]{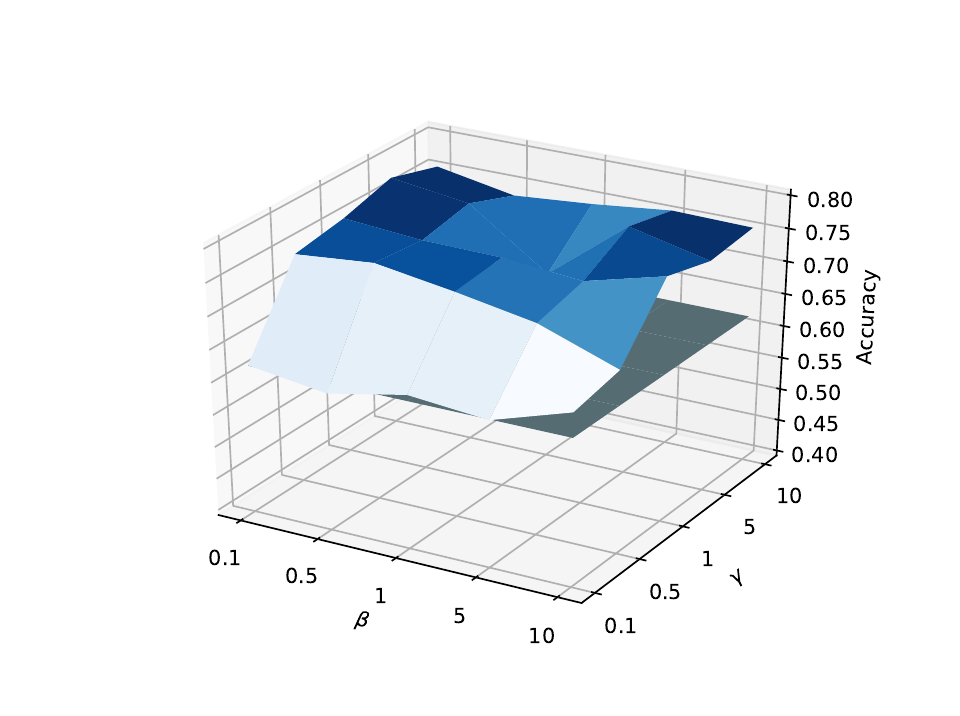}\label{3d_2}
\end{minipage}
\caption{This table illustrates the impact of FedMpae hyper-parameters $\beta$ and $\gamma$ on node classification for Cora (left) and Computer (right). The dark green plane is the accuracy of FedSage+.}
\label{3d}
\end{figure}

\paragraph{\textbf{Hyper-parameters in Loss Function.}}
Figure~\ref{3d} has presented the test accuracy with the combination of $\beta$ and $\gamma$. Our FedMpae plane is always on the FedSage+, which manifests that our model has significantly superior to fedsage+ in a low label rate. We set $\beta =1$ and $\gamma=1$ in this work.

All those hyper-parameter analyses manifest that our FedMpa is somewhat sensitive to various parameter combinations. Therefore, we need to fix some parameters to keep the model performance, which answers \textbf{Q2}.

\begin{figure}[htbp]
\centering
\begin{minipage}[c]{0.42\textwidth}
\centering
\includegraphics[width=\columnwidth]{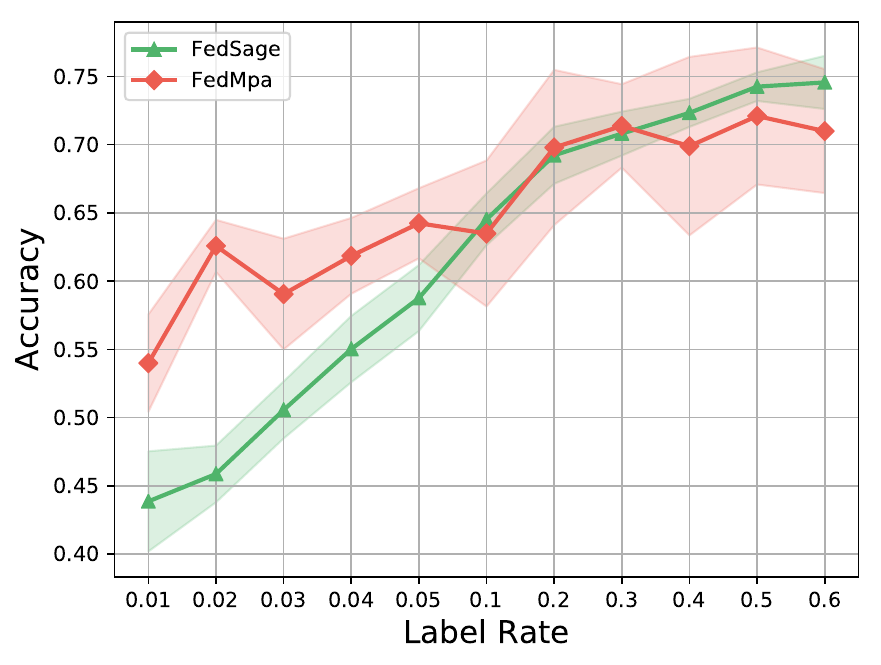}\label{figure3_3}
\end{minipage}
\hspace{0.1\textwidth}
\begin{minipage}[c]{0.42\textwidth}
\centering
\includegraphics[width=\columnwidth]{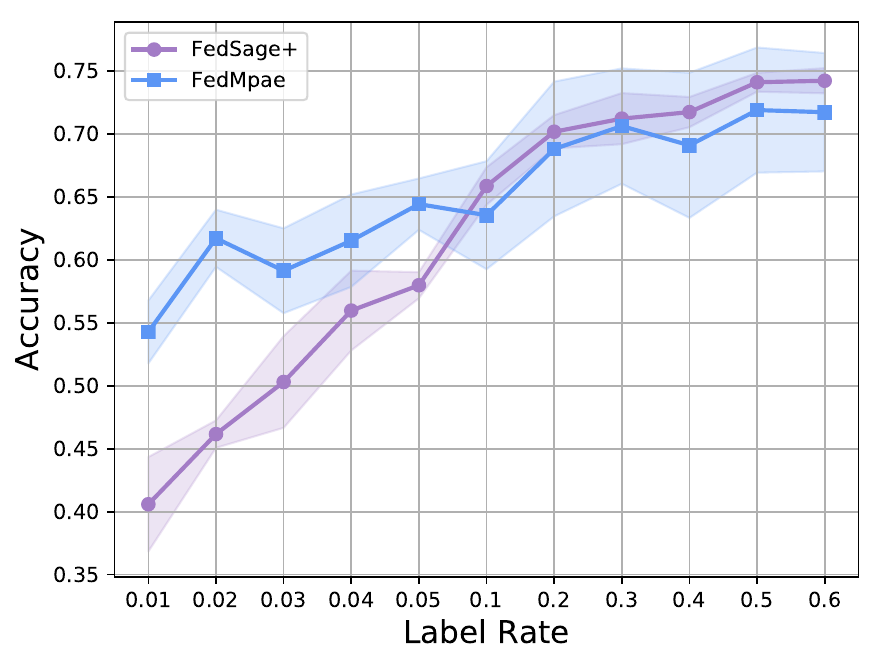}\label{figure3_4}
\end{minipage}
\caption{This table illustrates the impact of label rate on node classification for Citeseer. As is evident from the figure, the accuracy of both FedSage and FedSage+ increases with an increase in label rate. Moreover, FedMpa and FedMpae have a pronounced advantage when the label rate is low.}
\label{figure3}
\end{figure}

\paragraph{\textbf{Label Rate.}}
Figure~\ref{figure3} displays the performance of FedMpa and FedSage on the Citeseer dataset across varying label rates. Our results indicate that FedMpa outperforms FedSage when the label rate ranges from 1\% to 5\%. However, FedSage exhibits better performance with 5\%- label rates(our FedMpa still shows its competitive abilities). The chief reason is that the sampling track in FedSage can further promote the model generalization with the increased label rate. Thus, we conclude that FedMpa and FedSage complement each other's strengths, which answers \textbf{Q3}. FedMpa represents a continuation of the work on FedSage under low-label rate scenarios, highlighting its potential for use in real-world scenarios. 

\subsection{Analysis on Online Calculation}

\begin{figure}[htbp]
\centering
\begin{minipage}[c]{0.48\textwidth}
\centering
\includegraphics[width=\columnwidth]{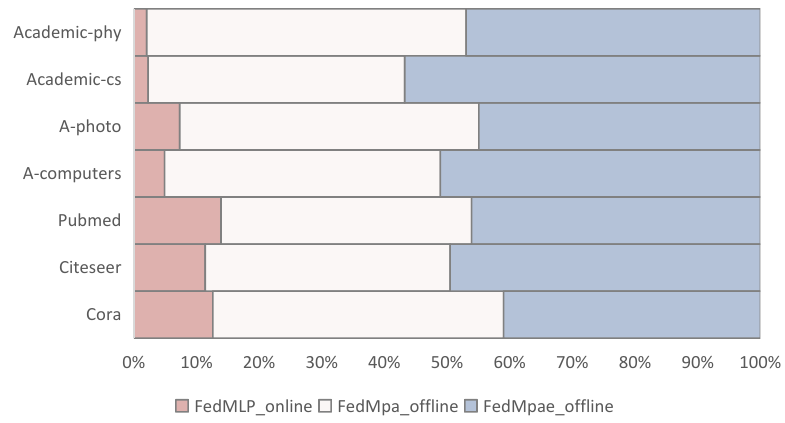}
\end{minipage}
\hspace{0.02\textwidth}
\begin{minipage}[c]{0.48\textwidth}
\centering
\includegraphics[width=\columnwidth]{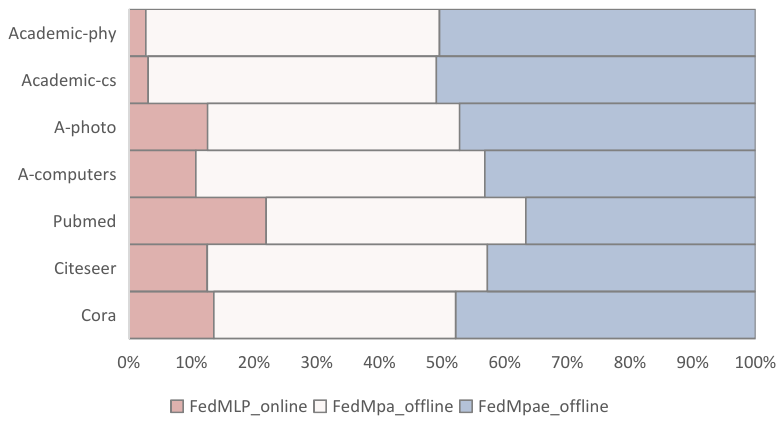}
\end{minipage}
\caption{Left: $M=3$ and label rate 1\%. Right: $M=3$ and label rate 5\%. The figure illustrates the rate of occupation for online and offline calculation time. The higher the rate, the more online calculation costs and communication rounds are required to arrive at an optimal solution.}
\label{figure4}
\end{figure}

As illustrated in Figure~\ref{figure4}, the online calculation time constitutes a minor fraction of the overall calculation time, highlighting the efficiency of our model in performing tasks with demanding online computation requirements. The reason lies in our online model's linear property, which requires fewer training epochs.

\section{Conclusions}
This work addresses the problem of training a local GNN model by setting a few labeled data and missing cross-subgraph edges at a federal level. To this end, we propose an SFL framework, FedMpa, which leverages global features to conduct local subgraph tasks. To cope with the missing edges, we introduce the FedMpae method to relearn the substructure with an innovation view that applies pooling operation to form super-nodes. We compared our FedMpa model with FedSage over six datasets with a low label rate, and our experiments achieved consistently competitive performance. In the future, we intend to design more federated graph models to suit a variety of related scenarios.
%
%
%
\bibliographystyle{splncs04}
\bibliography{mybibliography}
\end{document}